# Graph Regularized Nonnegative Matrix Factorization for Hyperspectral Data Unmixing

Roozbeh Rajabi, Mahdi Khodadadzadeh, Hassan Ghassemian
Faculty of Electrical and Computer Engineering
Tarbiat Modares University (TMU)
Tehran, Iran
{r.rajabi, khodadadzadeh, ghassemi}@modares.ac.ir

*Abstract*—Spectral unmixing is an important tool in hyperspectral data analysis for estimating endmembers and abundance fractions in a mixed pixel. This paper examines the applicability of a recently developed algorithm called graph regularized nonnegative matrix factorization (GNMF) for this aim. The proposed approach exploits the intrinsic geometrical structure of the data besides considering positivity and full additivity constraints. Simulated data based on the measured spectral signatures, is used for evaluating the proposed algorithm. Results in terms of abundance angle distance (AAD) and spectral angle distance (SAD) show that this method can effectively unmix hyperspectral data.

## I. INTRODUCTION

In hyperspectral imagery mixed pixels consist of more than one distinct material as illustrated in Fig.1. There are two reasons behind existence of mixed pixels. First reason is related to low spatial resolution of hyperspectral sensors. The second reason is combining different materials forming a homogenous mixture that is independent of spatial resolution of the sensors.

Spectral mixture analysis (or spectral unmixing) refers to decomposition of mixed pixels into endmembers and abundance fractions. Endmembers are extracted spectrum of distinct materials and abundance fractions are defined as the proportions of the extracted endmembers in mixed pixels [1].

Mixing can be modeled in two different ways: Linear and Nonlinear. In Linear Mixing Model (LMM) the measured spectrum is a linear combination of endmembers spectra added by observation noise error. Nonlinear mixing model is resulted in the intimate mixture of materials. Most of the unmixing methods are based on LMM [2].

Many methods based on LMM have been proposed in literature. These methods can be categorized as geometrical or statistical based algorithms [3]. Some examples of geometrical based methods are pixel purity index (PPI) [4], N-FINDR [5], vertex component analysis (VCA) [6], convex cone analysis (CCA) [7], nonnegative matrix factorization (NMF) [8]. Statistical methods like Bayesian analysis of spectral mixture data using Markov chain Monte Carlo methods [9] are based on Bayesian framework.

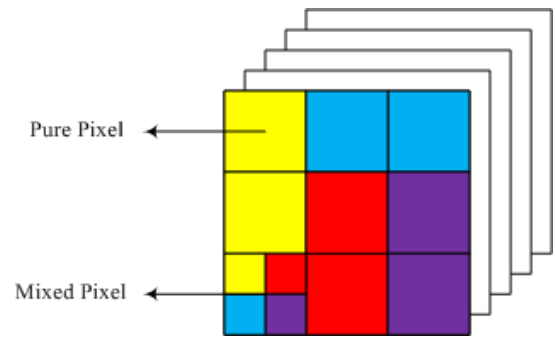

Fig. 1. Pure and Mixed Pixels in Hyperspectal Imagery.

There are also some semi-supervised methods for spectral unmixing. These methods search for the best collection of signatures within a relatively large known spectral library to optimally model each mixed pixel [3]. Sparse regression methods have been used for this purpose in literature [10].

In this paper graph regularized nonnegative matrix factorization (GNMF) has been examined for hyperspectral unmixing. The proposed method has been applied on simulated data that are generated using USGS spectral library [14]. The results show that this method performs better in comparison with NMF.

Section 2 presents a brief description of linear mixing model. In section 3 NMF and GNMF are discussed. Description of the database used for evaluating the proposed algorithm is provided in section 4. The simulation results are also presented in this section. Section 5 summarizes and concludes the paper.

## II. LINEAR SPECTRAL MIXTURE ANALYSIS

Linear spectral mixture model is a widely used model for spectral unmixing of hyperspectral data that many methods are based on it. Assume that L is the number of spectral bands. The measured spectrum (X) can be expressed by the equation (1).

$$X = SA + W \qquad (1)$$

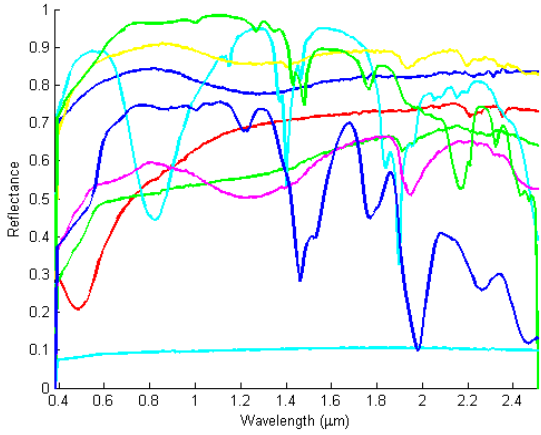

Fig. 2. Selected Materials from USGS Library.

In this equation S is spectral signature matrix of endmembers, A is abundance fraction matrix and W is an additive noise matrix. Each column of X ($x_n$) is a linear combination of spectral signatures in S as formulated in (2).

$$x_n = Sa + w = \sum_{i=1}^{P} a_i s_i + w$$
$$= a_1 s_1 + a_2 s_2 + .... + a_P s_P + w \ , \ 1 \leq n \leq N \quad (2)$$

In (2) $P$ is the number of endmembers, $N$ is the number of pixels and **a** is the vector of abundance fractions for $n$th pixel. There are two physical constraints on abundance fraction values that should be considered: Abundance non-negativity constraint (ANC) and abundance sun-to-one constraint (ASC) [2].

### III. METHODS

#### A. Nonnegative Matrix Factorization (NMF)

For a given matrix X, NMF finds nonnegative matrix factors U and V such that:

$$X \approx UV^T \quad (3)$$

For quantifying the quality of the approximation, cost functions based on Euclidean distance or Kullback-Leibler divergence can be used. Cost function for NMF using Euclidean distance is given in (4).

$$O = \left\| X - UV^T \right\|^2 \quad (4)$$

Minimizing this cost function with respect to U and V subjected to $U, V \geq 0$ will lead to NMF method [11].

#### B. Graph Regularized NMF (GNMF)

GNMF works based on local invariance assumption that assumes if two points are close in the intrinsic geometry of data, the representation of these points in the new basis are close to each other. GNMF uses geometrical based regularizer to preserve Riemannian structure.

To model the geometric structure of data, consider a nearest neighbor graph on a scatter of a data points. There is an edge between two data points ( $x_j$ , $x_l$ ) if they are neighbors. Different neighborhood systems are applicable like 4-neighbourhood or 8-neighbourhood systems. Weight Matrix (W) on the graph can be defined using many methods. The simplest one is 0-1 weighting. In this method if two pixels have an edge between them the weight will be 1 ($W_{jl} = 1$). Other weighting methods include heat kernel, dot product weighting and etc.

Using Euclidean distance, cost function for considering geometrical properties is defined by following equation.

$$R = \frac{1}{2} \sum_{j,l=1}^{N} \left\| z_j - z_l \right\|^2 W_{jl}$$
$$= \sum_{j=1}^{N} z_j^T z_j D_{jj} - \sum_{j,l=1}^{N} z_j^T z_l D_{jl} \quad (5)$$
$$= Tr(V^T DV) - Tr(V^T WV) = Tr(V^T LV)$$

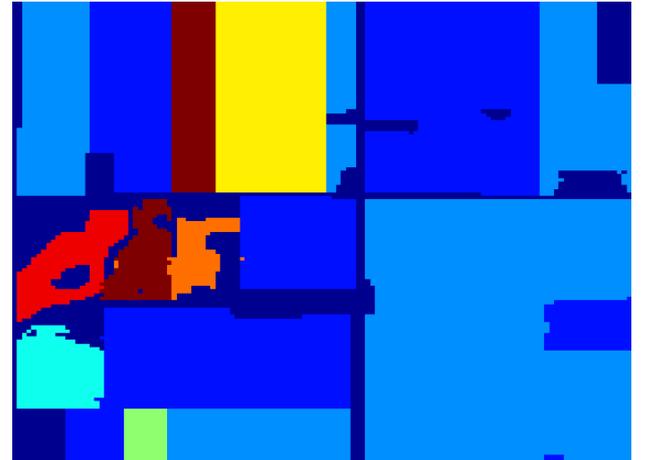

(a)

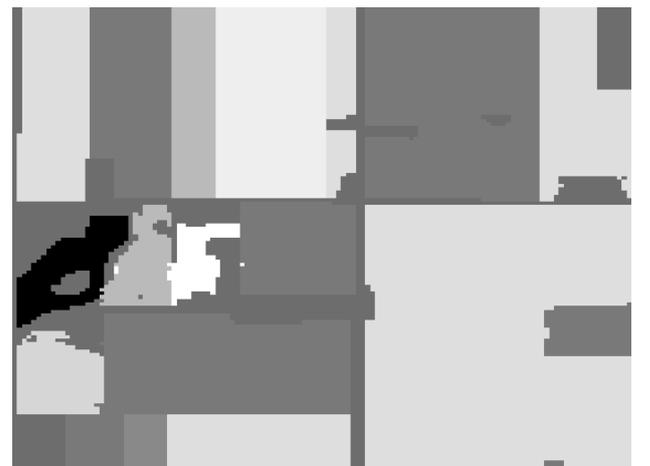

(b)

Fig. 3. F210 Sub Scene (a) Ground truth map of high resolution image. (b) Band 30 of high resolution simulated image.

In this equation $z_j$ is the low dimensional representation of $x_j$ (point in original basis). Tr(.) denotes the trace of a matrix. D and L are defined as follows:

$$D_{jj} = \sum_l W_{jl}$$
$$L = D - W \quad (6)$$

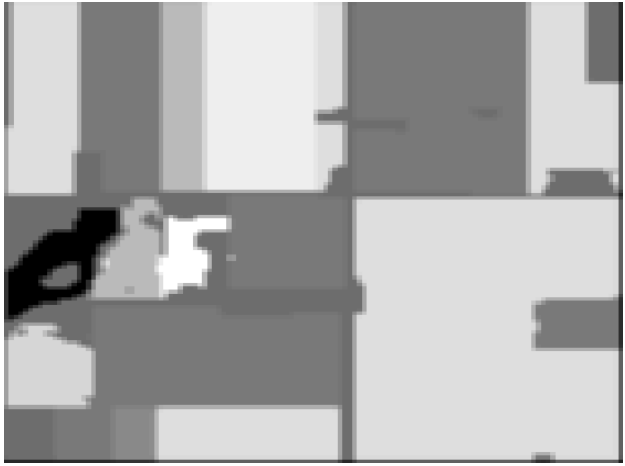
(a)

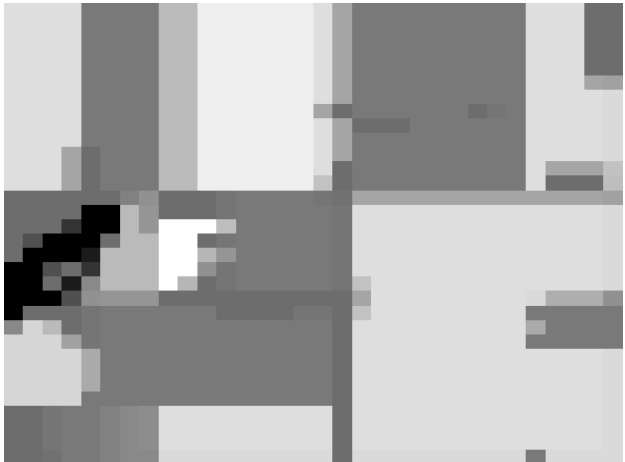
(b)

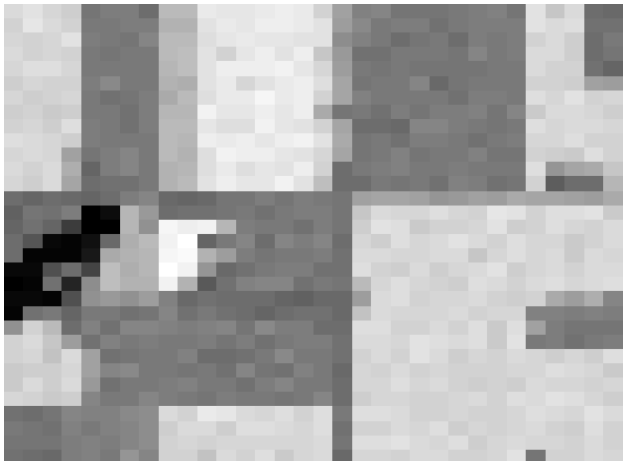
(c)

Fig. 4. Simulated Image (a) Band 30 of filtered high resolution image. (b) Band 30 of low resolution image. (c) Band 30 of low resolution data with added noise.

Combining cost function of (5) with original NMF cost function of (4) and minimizing that with respect to U and V will lead to GNMF method. Implemented algorithms for GNMF are available online[1] by the authors of [12]. Using Euclidean distance, the overall cost function is given by the equation (7) [12].

$$O_{Euclidean} = \|X - UV^T\|^2 + \lambda \text{Tr}(V^T L V) \quad (7)$$

$\lambda \geq 0$ is the regularization parameter and controls the smoothness of the new representation of data.

### C. Proposed Algorithm using GNMF

First number of endmembers should be determined. Dimension reduction methods like PCA or HySime [13] can be used for this aim. In this paper principal component analysis (PCA) is used for estimating the number of endmembers.

Adding abundance sum to one constraint (ASC) [8] subjected to GNMF method, the proposed method is given in the following algorithm.

**Algorithm:**

**Initialization:**
- Initializing U (N×k) using random nonnegative values
- Initializing V (k×L) using random nonnegative values
- Scale columns of V to sum to one.

**Loop:**
- Minimize the cost function in (7)
- Scale columns of V to sum to one.

**Until converging or stop criteria.**

### IV. EXPERIMENTS AND RESULTS

#### A. Simulated Data

USGS digital spectral library [14] has been used to simulate data for experiments. Four different materials have been chosen to generate simulated data. Fig. 2 shows spectral signatures of the selected materials.

To generate more realistic data, a sub scene of F210 multispectral data used for generating simulated data. Fig. 3 illustrates the ground truth and sample band of selected sub scene image. Measured spectrum of each pixel was substituted by signature of the selected materials. Then image was degraded by a scale factor of k (k=3) to simulate mixed pixels. For this purpose first the image was filtered by a k×k Gaussian filter. Then it was resized to form a low resolution image. Therefore the resulted image has mixed pixels with known abundance values and endmember signatures. Gaussian noise with SNR=30dB was added to simulate error noise (see Fig. 4) [15].

#### B. Results

NMF and GNMF methods have been applied on the simulated data and performance of proposed method has been evaluated using spectral angle distance (SAD) and abundance angle error (AAD).

---
[1] http://www.zjucadcg.cn/dengcai/

Considering $m_i$ and $\hat{m}_i$ as i-th original signature and estimated signature respectively, SAD measures the similarity between original signatures and estimated signatures and is given by the equation (8).

$$\text{SAD}_{m_i} = \cos^{-1}\left(\frac{m_i^T \hat{m}_i}{\|m_i\|\|\hat{m}_i\|}\right) \qquad (8)$$

Based on SAD, the root mean square error distance is given by the equation (9).

$$\text{rms}_{\text{SAD}} = \left(\frac{1}{p}\sum_{i=1}^{p}(\text{SAD}_{m_i})^2\right)^{1/2} \qquad (9)$$

Consider that $a_i$ and $\hat{a}_i$ are original abundance fractions and estimated abundance fractions of i-th pixel respectively, AAD measures the similarity between original abundance fractions and estimated abundance fractions and is given by the equation (10).

$$\text{AAD}_{a_i} = \cos^{-1}\left(\frac{a_i^T \hat{a}_i}{\|a_i\|\|\hat{a}_i\|}\right) \qquad (10)$$

Based on AAD, the root mean square error distance is given by equation (11) [16].

$$\text{rms}_{\text{AAD}} = \left(\frac{1}{N}\sum_{i=1}^{N}(\text{AAD}_{a_i})^2\right)^{1/2} \qquad (11)$$

TABLE I summarizes the performance evaluation results in terms of defined measures. Results show better performance of GNMF method in comparison with original NMF algorithm.

TABLE I: Performance evaluation results.

|  | NMF | GNMF |
|---|---|---|
| $\text{rms}_{\text{SAD}}$ (degrees) | 19.54 | 15.76 |
| $\text{rms}_{\text{AAD}}$ (degrees) | 10.23 | 8.34 |

## I. Conclusion

Spectral unmixing for hyperspectral imagery is an emerging field of study in remote sensing applications. Mixed pixels exist in hyperspectral data mostly because of relatively low spatial resolution of hyperspectral sensors. Linear mixing model (LMM) was used for modeling the mixing procedure.

In this paper graph regularized nonnegative matrix factorization method (GNMF) has been used for spectral mixture analysis of hyperspectral imagery. The motivation of using this method is exploiting geometrical structure of the data by GNMF. Results in terms of SAD and AAD show better performance of this method in comparison with original NMF method.

In future work, different cost function based on KL divergence and other methods for generating weight matrix can be examined. For further experiments and evaluation, the algorithm should be applied on real datasets.

## Acknowledgment

The authors would like to express their sincere thanks to Dr. D. Cai for his valuable help.

## References

[1] N. Keshava and J. F. Mustard, "Spectral Unmixing," *IEEE Signal Processing Magazine,* vol. 19, pp. 44-57, 2002.
[2] N. Keshava, "A Survey of Spectral Unmixing Algorithms," *Lincoln Lab Journal,* vol. 14, pp. 55-78, 2003.
[3] J. Bioucas-Dias and A. Plaza, "An Overview on hyperspectral unmixing: geometrical, statistical, and sparse regression based approaches," in *Geoscience and Remote Sensing Symposium,2011 IEEE International,IGARSS 2011*, Canada, 2011.
[4] J. W. Boardman, F. A. Kruse, and R. O. Green, "Mapping target signatures via partial unmixing of AVIRIS data," in *Summaries of JPL Airborne Earth Science Workshop*, 1995, pp. 95-1.
[5] M. E. Winter, "N-FINDR: an algorithm for fast autonomous spectral end-member determination in hyperspectral data," in *SPIE conference on Imaging Spectrometry V*, 1999, pp. 266-275.
[6] J. M. P. Nascimento and J. M. B. Dias, "Vertex component analysis: a fast algorithm to unmix hyperspectral data," *Geoscience and Remote Sensing, IEEE Transactions on,* vol. 43, pp. 898-910, 2005.
[7] A. Ifarraguerri and C. I. Chang, "Multispectral and hyperspectral image analysis with convex cones," *Geoscience and Remote Sensing, IEEE Transactions on,* vol. 37, pp. 756-770, 1999.
[8] V. P. Pauca, J. Piper, and R. J. Plemmons, "Nonnegative matrix factorization for spectral data analysis," *Linear Algebra and its Applications,* vol. 416, pp. 29-47, 2006.
[9] S. Moussaoui, C. Carteret, D. Brie, and A. Mohammad-Djafari, "Bayesian analysis of spectral mixture data using Markov chain Monte Carlo methods," *Chemometrics and Intelligent Laboratory Systems,* vol. 81, pp. 137-148, 2006.
[10] M. D. Iordache, J. M. Bioucas-Dias, and A. Plaza, "Sparse Unmixing of Hyperspectral Data," *Geoscience and Remote Sensing, IEEE Transactions on,* vol. 49, pp. 2014-2039, 2011.
[11] D. D. Lee and H. S. Seung, "Algorithms for non-negative matrix factorization," *Advances in neural information processing systems,* vol. 13, 2000.
[12] D. Cai, X. He, J. Han, and T. S. Huang, "Graph Regularized Nonnegative Matrix Factorization for Data Representation," *Pattern Analysis and Machine Intelligence, IEEE Transactions on,* vol. 33, pp. 1548-1560, 2011.
[13] J. M. P. Nascimento and J. M. Bioucas-Dias, "Hyperspectral signal subspace estimation," in *Geoscience and Remote Sensing Symposium, 2007. IGARSS 2007. IEEE International*, 2007, pp. 3225-3228.
[14] R. N. Clark, G. A. Swayze, R. Wise, E. Livo, T. Hoefen, R. Kokaly, and S. J. Sutley. USGS digital spectral library splib06a [Online]. Available: http://speclab.cr.usgs.gov/spectral.lib06
[15] A. Villa, J. Chanussot, J. A. Benediktsson, and C. Jutten, "Spectral Unmixing for the Classification of Hyperspectral Images at a Finer Spatial Resolution," *Selected Topics in Signal Processing, IEEE Journal of,* vol. 5, pp. 521-533, 2011.
[16] J. M. P. Nascimento, "Unsupervised Hyperspectral Unmixing," Universidade Tecnica de Lisboa, 2006.